%% file: main.tex
\documentclass{article}

\usepackage{microtype}
\usepackage{graphicx}
\usepackage{float}
\usepackage{subfigure}
\usepackage{booktabs}
\usepackage{float}
\usepackage{dsfont}
\usepackage{tabularx}
\usepackage{highlight}
\usepackage{collcell}
\usepackage{multirow}
\usepackage[export]{adjustbox}
\usepackage{amsmath,amssymb,amsthm}
\usepackage{xcolor}

\usepackage{pgf}
\usepackage{collcell}
\usepackage{booktabs}
\usepackage{tabularx}
\usepackage{url}
\usepackage{graphicx}
\usepackage{graphics}
\usepackage{xstring}
\usepackage{hyperref}
\usepackage{inconsolata}

\usepackage[accepted]{icml2021}

\icmltitlerunning{Factors of Influence of the Overestimation Bias of Q-Learning}

\begin{document}

\twocolumn[
\icmltitle{Factors of Influence of the Overestimation Bias of Q-Learning}




\begin{icmlauthorlist}
\icmlauthor{Julius Wagenbach}{to}
\icmlauthor{Matthia Sabatelli}{to}

\end{icmlauthorlist}

\icmlaffiliation{to}{Bernoulli Institute for Mathematics, Computer Science and Artificial Intelligence University of Groningen, The Netherlands}

\icmlcorrespondingauthor{Matthia Sabatelli}{m.sabatelli@rug.nl}

\icmlkeywords{Machine Learning, ICML}

\vskip 0.3in
]



\printAffiliationsAndNotice{\icmlEqualContribution} 

\begin{abstract}

We study whether the learning rate $\alpha$, the discount factor $\gamma$ and the reward signal $r$ have an influence on the overestimation bias of the Q-Learning algorithm. Our preliminary results in environments which are stochastic and that require the use of neural networks as function approximators, show that all three parameters influence overestimation significantly. By carefully tuning $\alpha$ and $\gamma$, and by using an exponential moving average of $r$ in Q-Learning's temporal difference target, we show that the algorithm can learn value estimates that are more accurate than the ones of several other popular model-free methods that have addressed its overestimation bias in the past. 

\end{abstract}

\section{Introduction and Preliminaries}
Reinforcement Learning (RL) is a machine learning paradigm that aims to train agents such that they can interact with an environment and maximize a numerical reward signal. While there exist numerous ways of learning from interaction, in model-free RL this is achieved by learning value functions that estimate how good it is for an agent to be in a certain state, or how good it is for the agent to take a certain action in a particular state \cite{sutton2018reinforcement}. The goodness/badness of state-action pairs is typically expressed in terms of expected future rewards: the higher the expected value of a state-action tuple, the better it is for the RL agent to perform a certain action in a given state. Estimating state-action values accurately is therefore key when it comes to model-free RL, as it is in fact the agent's value functions that define its actions and, as a result, allow it to interact optimally with its environment.  

To express such concepts more formally, let us define the RL setting as a Markov Decision Process (MDP) represented by the following tuple $(\mathcal{S}, \mathcal{A}, P, r, \gamma)$ \cite{puterman2014markov}. Its components are a state space $\mathcal{S}$, an action space $\mathcal{A}$, a transition probability distribution $P$, that defines the probability for an agent to visit state $s$ given action $a$ at time step $t$, $p(s_{t+1} | s_{t}, a_{t})$, a reward signal $r$, coming from the reward function $\Re (s_{t}, a_{t}, s_{t+1})$, and a discount factor $\gamma \in [0, 1)$. The actions of the RL agent are selected based on its policy $\pi:\mathcal{S} \rightarrow \mathcal{A}$ that maps each state to an action.
For every state $s \in \mathcal{S}$, under policy $\pi$ the agent's state-value function $V^{\pi}: \mathcal{S} \rightarrow \mathds{R}$ is defined as:
\begin{align}
    V^{\pi}(s)=\mathds{E}\bigg[\sum_{k=0}^{\infty}\gamma^{k}r_{t+k}\bigg| s_t = s, \pi \bigg],
\label{eq:v}
\end{align}
while its state-action value function $Q: \mathcal{S}\times \mathcal{A} \rightarrow \mathds{R}
$ is defined as:
\begin{align}
    Q^{\pi}(s,a)=\mathds{E}\bigg[\sum_{k=0}^{\infty}\gamma^{k}r_{t+k} \bigg| s_t = s, a_t=a, \pi\bigg].
\end{align}
The main goal for the agent is to find a policy $\pi^{*}$ that realizes the optimal expected return:
\begin{align}
 V^{*}(s)=\underset{\pi}{\max}\:V^{\pi}(s), \ \text{for all} \ s\in\mathcal{S}
\end{align}
and the optimal $Q$ value function:
\begin{align}
Q^{*}(s,a)= \underset{\pi}{\max}\:Q^{\pi}(s,a) \ \text{for all} \ s\in\mathcal{S} \ \text{and} \ a \in\mathcal{A}.
\end{align} 

Learning these value functions is a well-studied problem in RL \cite{szepesvari2010algorithms, sutton2018reinforcement}, and several algorithms have been proposed to do so. The arguably most popular one is Q-Learning \cite{watkins1992q} which keeps track of an estimate of the optimal state-action value function $Q: \mathcal{S} \times \mathcal{A} \rightarrow \mathds{R}
$ and given a RL trajectory $\langle s_t, a_t, r_t, s_{t+1} \rangle$ updates $Q(s_t, a_t)$ with respect to the greedy target $r_t + \gamma \max_{a\in \mathcal{A}} Q(s_{t+1},a)$. Despite guaranteeing convergence to $Q^{*}(s,a)$ with probability $1$, Q-Learning is characterized by some biases that can prevent the agent from learning \cite{thrun1993issues, hasselt2010double,lu2018non}.

\subsection{The Overestimation Bias of Q-Learning}
Among such biases, the arguably most studied one is its overestimation bias: due to the maximization operator in its Temporal Difference (TD) target $\max_{a\in \mathcal{A}}Q(s_{t+1},a)$, Q-Learning estimates the expected maximum value of a state, instead of its maximum expected value, an issue which as discussed by \citet{van2011insights} dates back to research in order statistics \cite{clark1961greatest}. As a result, most recent work aimed to reduce Q-Learning's overestimation bias by replacing its $\max$ operator: Maxmin Q-Learning \cite{lan2020maxmin} controls it by trying to reduce the estimated variance of the different state-action values; whereas Variation Resistant Q-Learning \cite{pentaliotis2021variation} does so by keeping track of past state-action value estimates that can then be used when constructing the TD-target of the algorithm. On a similar note, \citet{karimpanal2021balanced} also define a novel TD-target which is a convex combination of a pessimistic and an optimistic term. A relatively older approach is that of \citet{hasselt2010double} who introduced the double estimator approach where one estimator is used for choosing the maximizing action while the other is used for determining its value. This approach plays a central role in his Double Q-Learning algorithm as well as in the more recent Weighted Double Q-Learning \cite{zhang2017weighted}, Double Delayed Q-Learning \cite{abed2018double} and Self-Correcting Q-Learning algorithms \cite{zhu2021self}. The recent rise of Deep Reinforcement Learning has shown that the overestimation bias of Q-Learning plays an even more important role when model-free RL algorithms are combined with deep neural networks, which in turn has resulted in a large body of works that have studied this phenomenon outside the tabular RL setting \cite{van2016deep, fujimoto2018addressing, kim2019deepmellow, cini2020deep, sabatelli2020deep, peer2021ensemble}.

\section{Methods}
\begin{figure*}[htb!]
  \centering
   \includegraphics[width=5cm,height=\textheight,keepaspectratio]{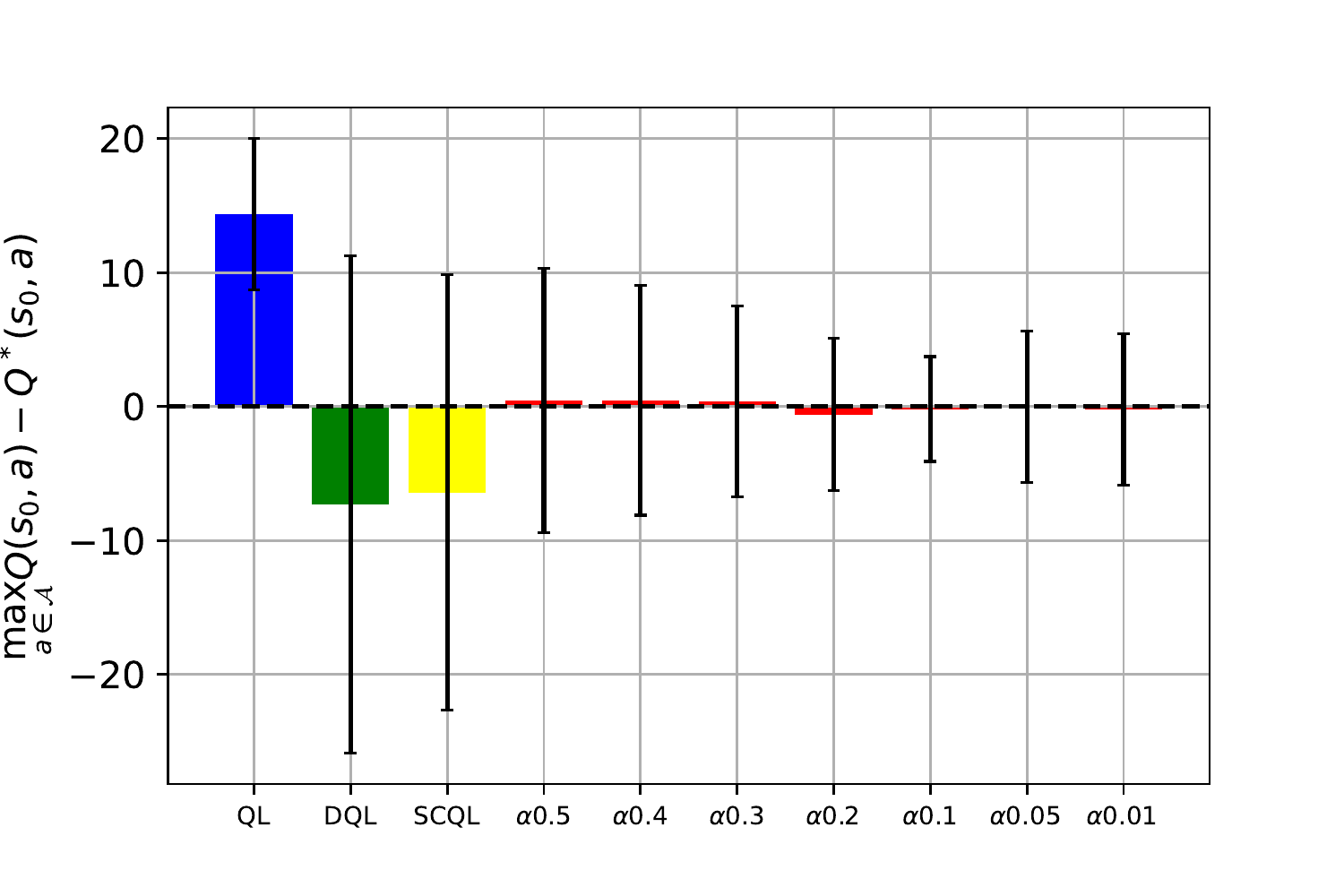}%
  \includegraphics[width=5cm,height=\textheight,keepaspectratio]{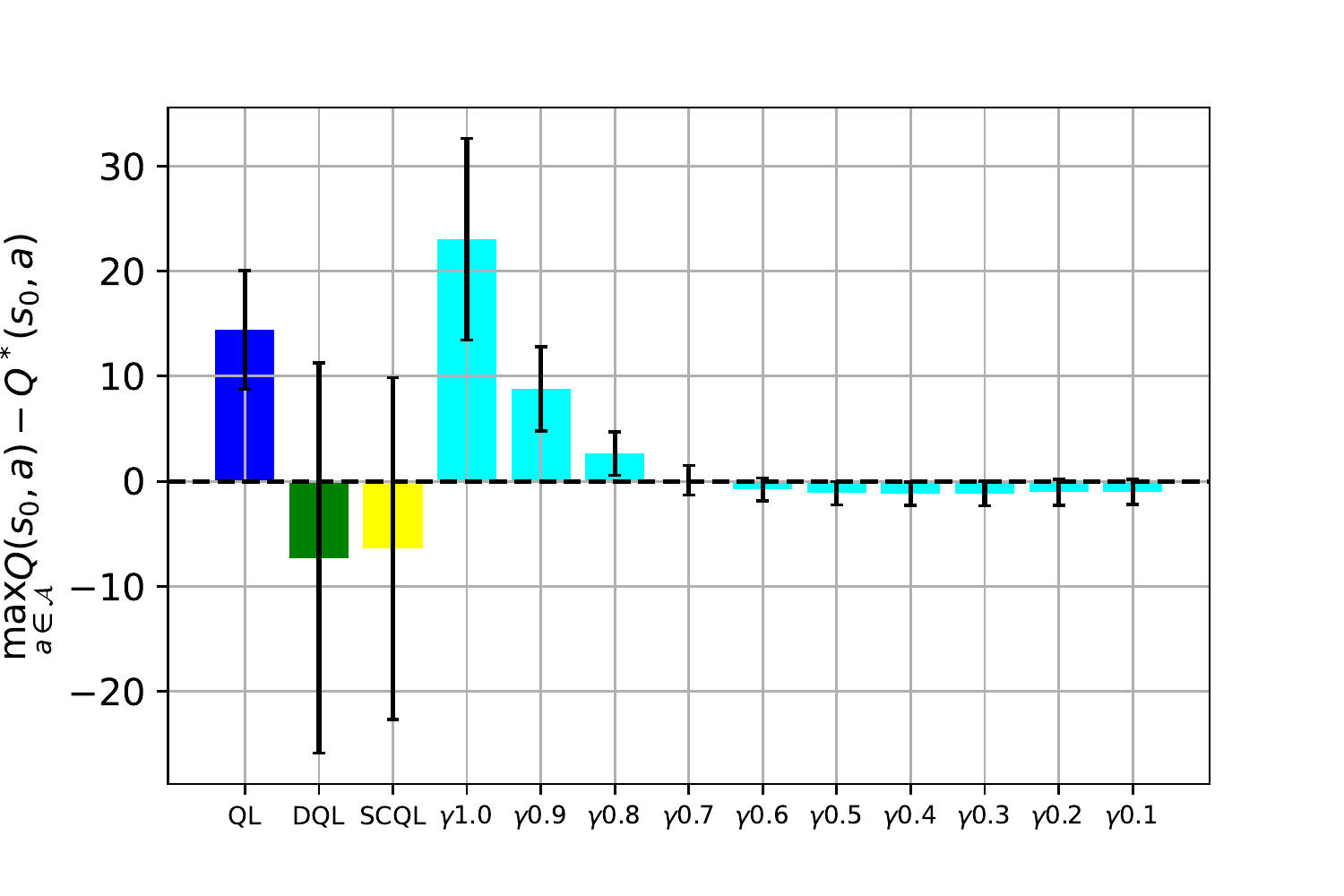}%
  \includegraphics[width=5cm,height=\textheight,keepaspectratio]{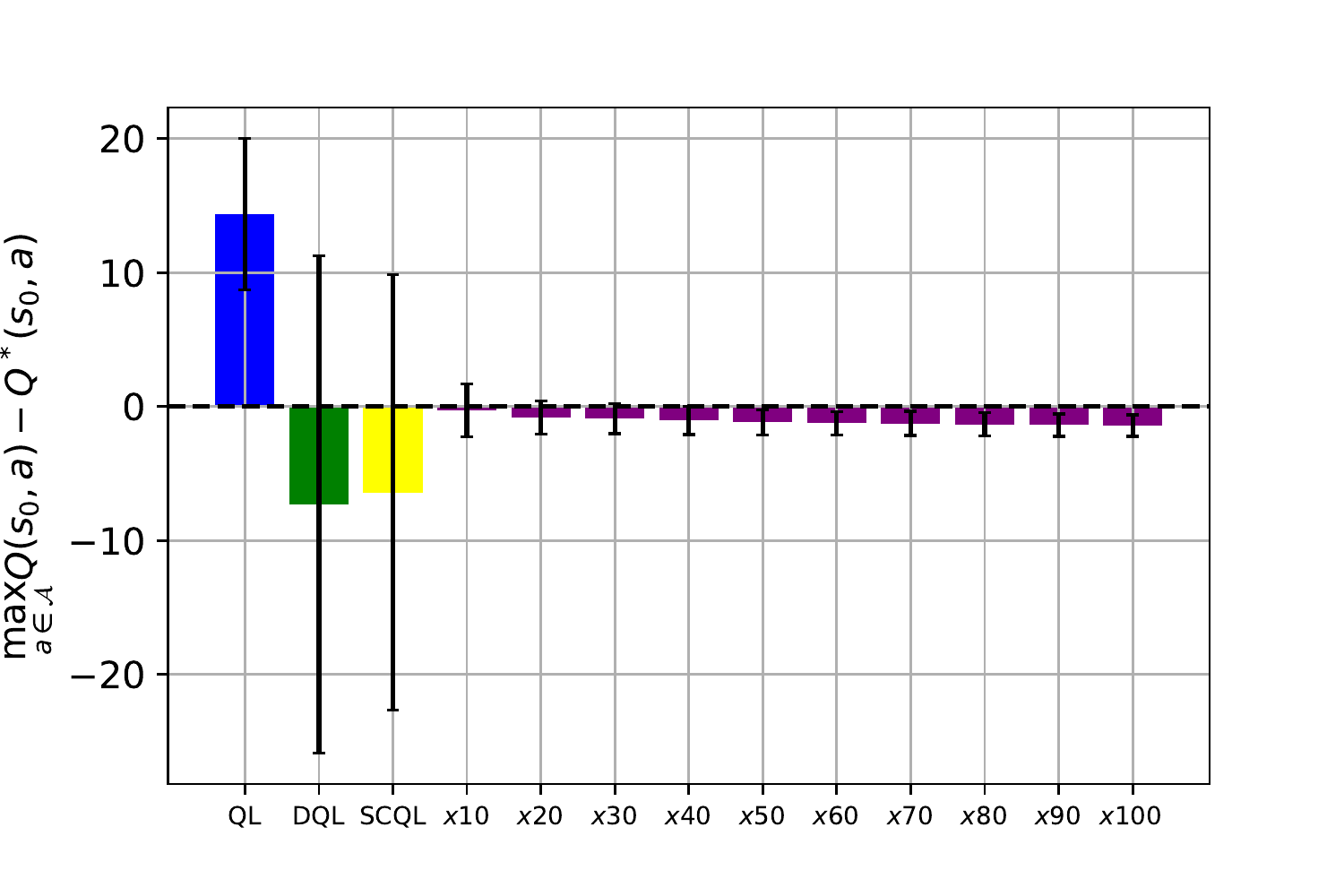}%
\caption{From left to right: our results showing that on the gridworld environment different values of $\alpha$, $\gamma$ and $x$ have an impact on the overestimation bias of Q-Learning. All factors of influence are compared against regular Q-Learning (blue), Double Q-Learning (green) and Self-Correcting Q-Learning (yellow).}
\label{fig:overview}
\end{figure*}
While, as explained earlier, reducing Q-Learning's overestimation bias has been mainly done by focusing on its $\max$ operator, in this paper, we take a different approach. Before introducing it, let us recall that Q-Learning learns $Q^{*}(s,a)$ as follows:
\begin{multline}
    Q(s_t,a_t):=Q(s_t,a_t) + \alpha\big[r_t + \gamma \:\underset{a\in \mathcal{A}}{\max} \: Q(s_{t+1},a) - \\ Q(s_t, a_t) \big].
\label{eq:q_learning}
\end{multline}

\subsection{Factors of Influence}
Instead of replacing the maximization estimator, we investigate whether overestimation can be prevented by tuning the following parameters: 
\begin{enumerate}
    \item \textbf{The learning rate $\alpha$} also denoted as the step-size parameter, controls the extent to which a certain state-action tuple gets updated with respect to the TD-target. Typically, small values imply slow convergence, while larger values may lead to divergence \cite{pirotta2013adaptive}. It is well known that Q-Learning's maximization estimator enhances the divergence of the algorithm, therefore we investigate whether its overestimation can be controlled by adopting learning rates which are low and fixed instead of linearly or exponentially decaying as done by \citet{hasselt2010double}.
    \item \textbf{The discount factor $\gamma$} enables to control the trade-off between immediate and long term rewards. While for many years it was considered best practice to set $\gamma$ to a constant value as close as possible to $1$, more recent research has demonstrated that this might not always be the best approach \cite{van2019using}. In fact a constant value of $\gamma$ has proven to yield time-inconsistent behaviours \cite{lattimore2014general}, failures in modelling agent's preferences \cite{pitis2019rethinking} and sub-optimal exploration \cite{franccois2015discount}. As mentioned by \citet{fedus2019hyperbolic} there seems to be a growing tension between the original $\gamma$ formulation and current RL research, which, however, has not yet been studied from an overestimation bias perspective, a limitation which we start addressing in this work.
    \item \textbf{The reward signal $r_t$} causes overestimation in environments where rewards are stochastic. The larger the variance in stochastic rewards, the higher the potential for overoptimistic values to accumulate and propagate through the system. However, if one averages the reward observed for a certain state-action pair over time, these averaged values would deviate from the true mean with a smaller variance. Therefore, we examine if overestimation can be reduced by using an exponential moving average in Q-Learning's TD-target which is computed as follows 
    \begin{equation}
        \hat{r}(s) \mathrel{+}= \frac{1}{x}(r(t)-\hat{r}(s)), 
    \end{equation}
    where $x$ is a static hyperparameter determining the degree of weighting decrease.
\end{enumerate}

\subsection{Experimental Setup}
We examine the effect on overestimation and performance of keeping $\alpha$ low and static, lowering $\gamma$, and using and averaged reward signal $\hat{r}$ instead of $r_t$ in three different environments, and compare the performance of Q-Learning (QL) to that of Double Q-Learning \cite{hasselt2010double} (DQL) and Self-Correcting Q-Learning \cite{zhu2021self} (SCQL). For the tabular setting, we use the Gridworld environment initially proposed by \citet{hasselt2010double}. The environment is a $3\times3$ grid with stochastic rewards in non-terminal states of a Bernoulli distribution $r\in [-12,10]$ and a fixed reward of 5 in the terminal state. We also test the effect of $\alpha,\gamma$ and $\hat{r}$ on the OpenAI gym \cite{brockman2016openai} \texttt{Blackjack-v0} environment, which simulates Blackjack including its stochastic state transitions and stochastic rewards. Lastly, we consider the function approximator case where we train a Deep-Q Network \cite{mnih2015human} on the \texttt{CartPole-v0} environment. In this environment, the value of the learning rate, and whether it is static or decaying, has already been shown to have a significant effect on overestimation \cite{chen2021investigation}. Our contribution here is examining the effect of the discount factor on overestimation. Note that, as rewards are deterministic, the method of replacing $r_t$ with $\hat{r}$ is not relevant here. We assess the impact of the aforementioned factors of influence by measuring the average gained reward obtained by the agent throughout training, as well as by quantifying the degree of overestimation by comparing $\max_{a\in \mathcal{A}}Q(s_0,a)$ to $Q^*(s_0,a)$. The latter is reported for all environments in Table \ref{tab:results} where we compute $\max_{a\in \mathcal{A}}Q(s_0,a) - Q^*(s_0,a)$: negative values mean that the tested algorithms suffer from underestimation bias, whereas positive values mean overestimation.

\section{Results}

\paragraph{Gridworld}

\begin{figure}[ht!]
  \centering
   \includegraphics[width=4.5cm,height=\textheight,keepaspectratio]{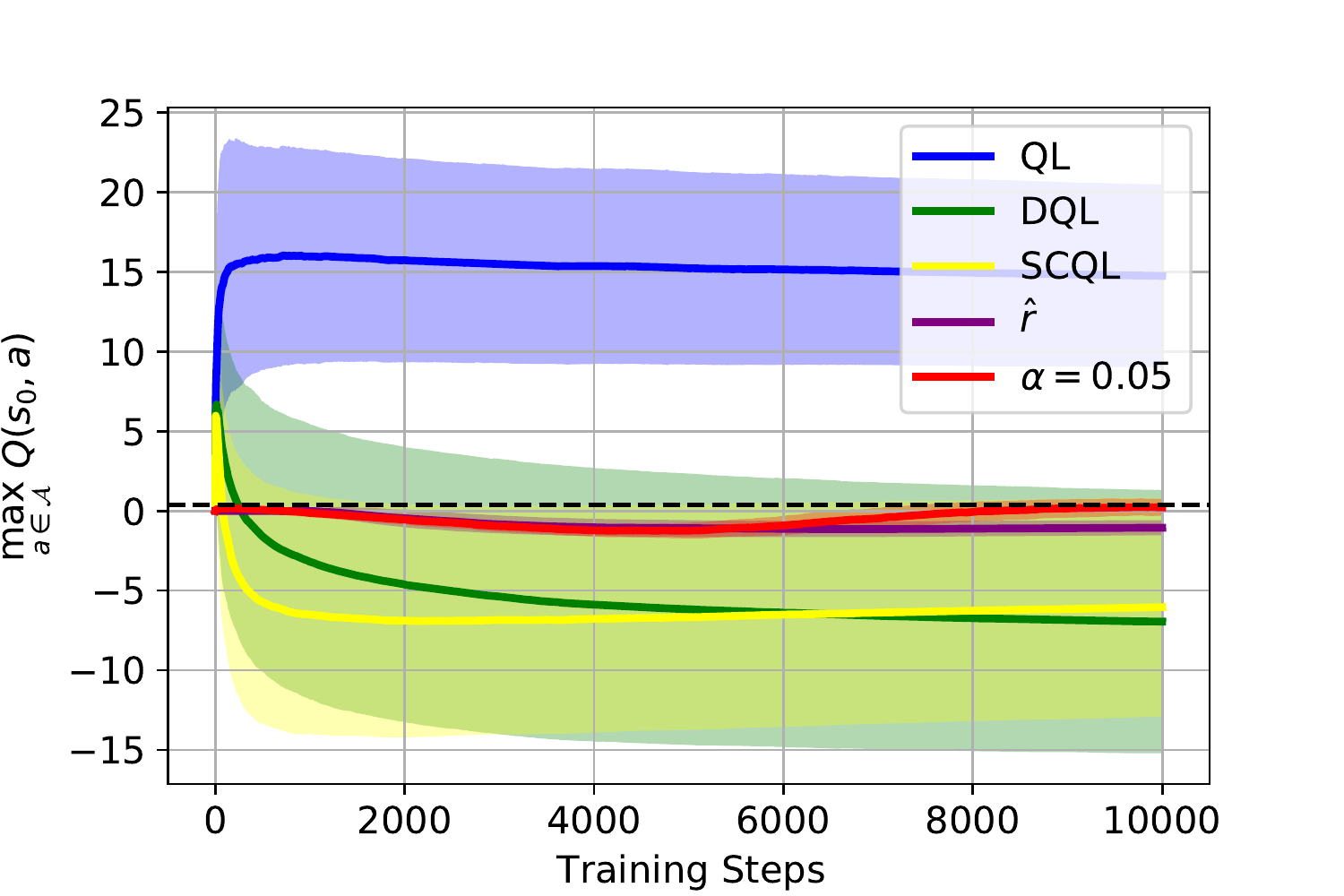}%
      \includegraphics[width=4.5cm,height=\textheight,keepaspectratio]{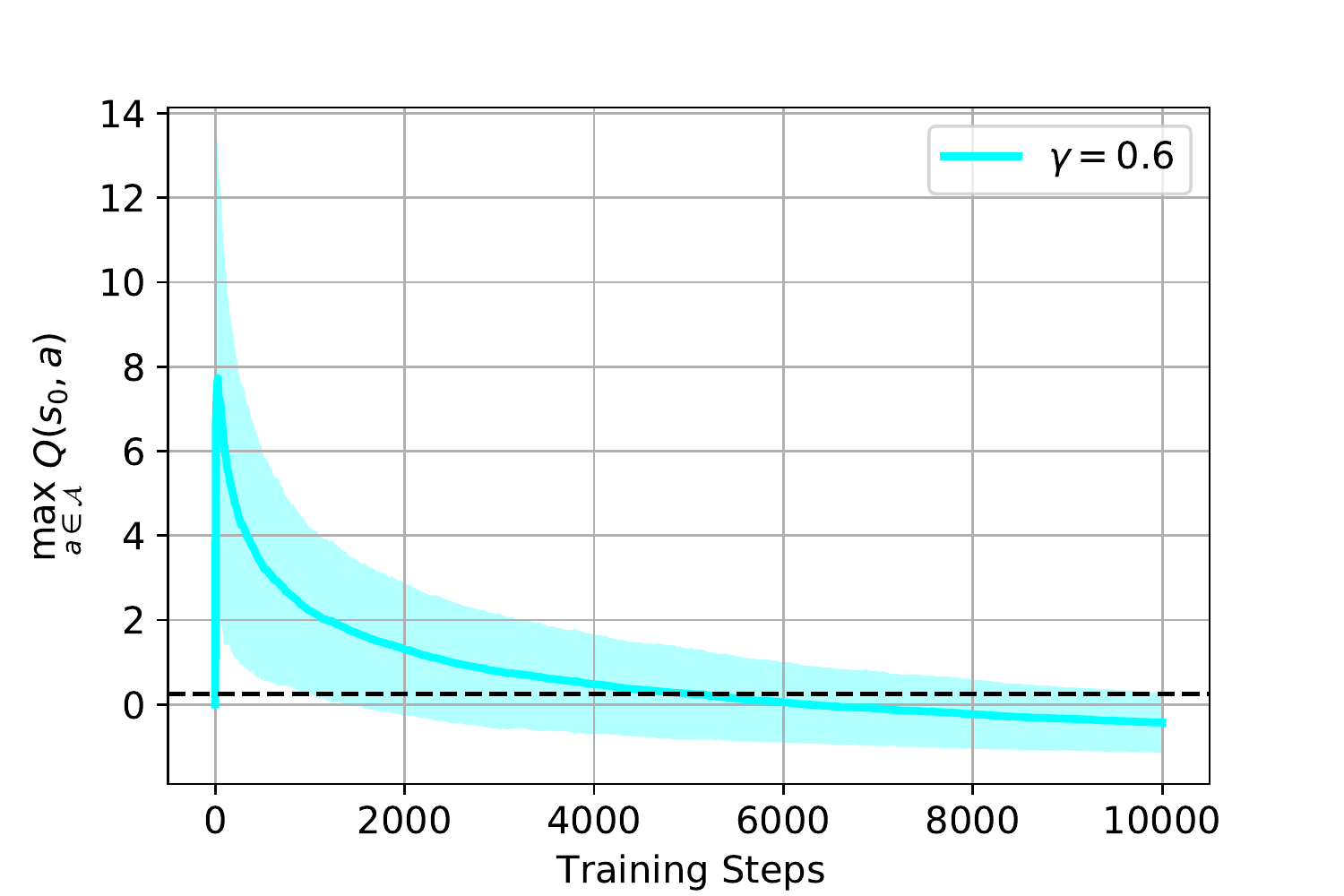}
      
  \includegraphics[width=4.5cm,height=\textheight,keepaspectratio]{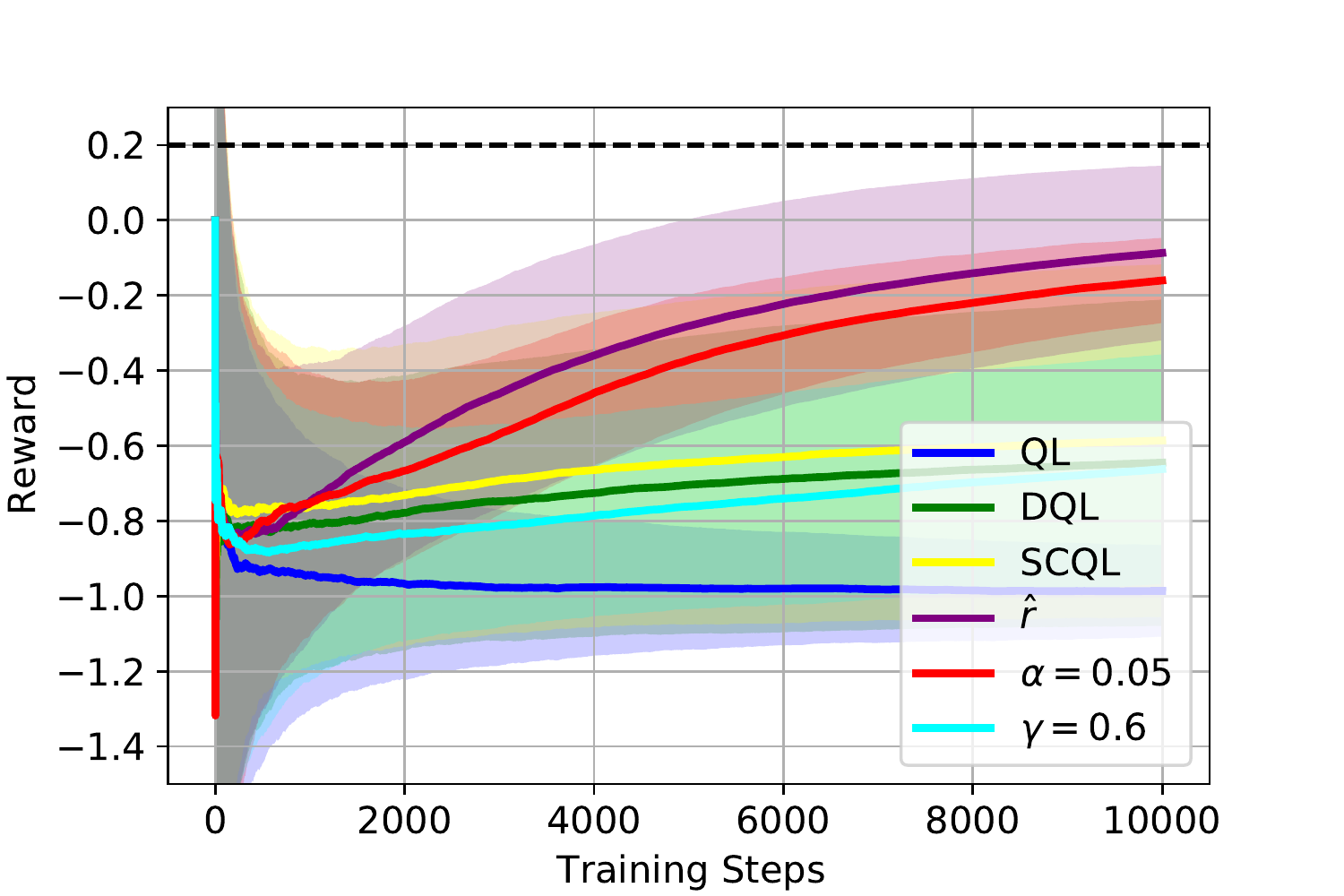}%
\caption{In clockwise order, our results showing that i) overestimation, as well as underestimation, can be prevented by using a constant value of $\alpha=0.05$ and by maintaining an exponential moving average estimate $\hat{r}$; ii) $\gamma=0.6$ allows Q-Learning to not overestimate; and iii) QL trained with either $\alpha=0.05$, $\gamma=0.6$ or $x=70$ outperforms regular QL, DQL and SCQL.}
\label{fig:gridworld_results}
\end{figure}

In the Gridworld environment, the optimal policy is to reach the goal state in 4 steps. Considering the stochastic rewards $r\in [-12,10]$ in non-terminal states, the correct value for $\max_{a\in \mathcal{A}}Q(s_{0},a)$ under $\pi^*$ for $\gamma = 0.95$ is $5 \gamma ^4 + \sum_{k=0}^{3} (-1)\gamma^k \approx 0.36$ (see black line in the first plot of Fig.\ref{fig:gridworld_results}). In the same figure we can see that QL trained with a dynamic $\alpha$ (blue line) produces an average estimate of $\approx 14.75$, therefore severely suffering from overestimation, whereas DQL and SCQL suffer from underestimation bias as they learn an estimate of $\approx -7.58$ and $-6.68$ respectively (see green and yellow lines). In Fig.\ref{fig:overview} we show what happens when experimenting with different values for $\alpha$ and $\gamma$ and, when computing $\hat{r}$ with different values of $x$. We can clearly see that all three parameters have a significant impact on Q-Learning's overestimation bias. Different values of $\alpha$ and $x$ result in very minor overestimation and underestimation (first and third plots of Fig.\ref{fig:overview}), whereas the lower the value of $\gamma$ (see second plot of Fig.\ref{fig:overview}), the less Q-Learning overestimates. Note however that lower values of $\gamma$ might not always allow Q-Learning to learn the optimal policy. For our experiments we found that best results were obtained with setting $\gamma$ to $0.6$ (second plot of Fig.\ref{fig:gridworld_results}). When it comes to $\alpha$ and $x$, Q-Learning estimates and performs best with $\alpha=0.05$ and $x=70$ (first and third plot of Fig.\ref{fig:gridworld_results}). As mentioned earlier, all factors of influence reduce Q-learning's overestimation bias, as summarized in Table \ref{tab:results}, while also increasing the obtained reward significantly. 

\begin{figure}[htb!]
  \centering
   \includegraphics[width=6cm,height=\textheight,keepaspectratio]{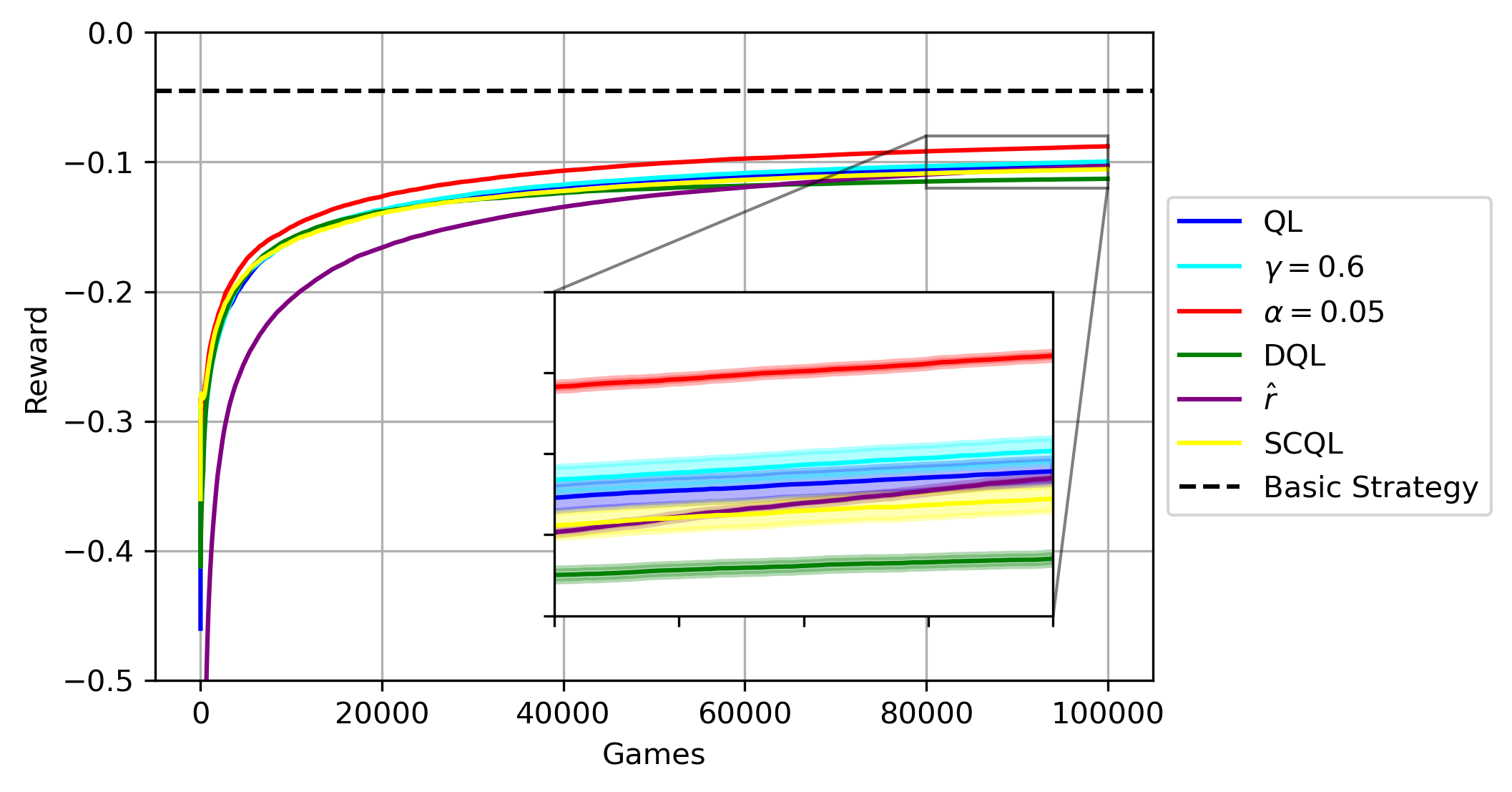}
  \caption{The rewards obtained by all algorithms on OpenAI's \texttt{Blackjack-v0} environment.}
\label{fig:blackjack}
\end{figure}

\paragraph{Blackjack} In Blackjack the optimal policy is known as basic strategy \citep{baldwin1956optimum} and results in an average reward of $-0.045$. We determine the degree of overestimation of the different algorithms by setting $\gamma=1$ and by subtracting the average of the starting state Q values weighted by the occurrence, by the optimal performance. For this experiment we only consider $\gamma=1$ as the number of steps it takes to finish an episode and receive the reward is irrelevant to the strategy. We again observe that DQL and SCQL suffer from underestimation bias (see second row of Table \ref{tab:results}) which, as shown by the reward plot in Fig.\ref{fig:blackjack}, results in suboptimal performance. Because these algorithms underestimate state-action values, they tend to pursue a more defensive strategy that makes the agent stand its hand instead of drawing a new card. 

A static $\alpha$ and $\hat{r}$ slightly result in overestimation, which however does not prevent the algorithms from outperforming DQL and SCQL (see Fig.\ref{fig:blackjack}). Surprisingly, also $\gamma=0.6$ results in better performance, however for this experiment we do not compute the degree of overestimation because of the complexity of the state space.  

\paragraph{Deep Q-Learning}

\begin{figure}[htb!]
  \centering
   \includegraphics[width=4.5cm,height=\textheight,keepaspectratio]{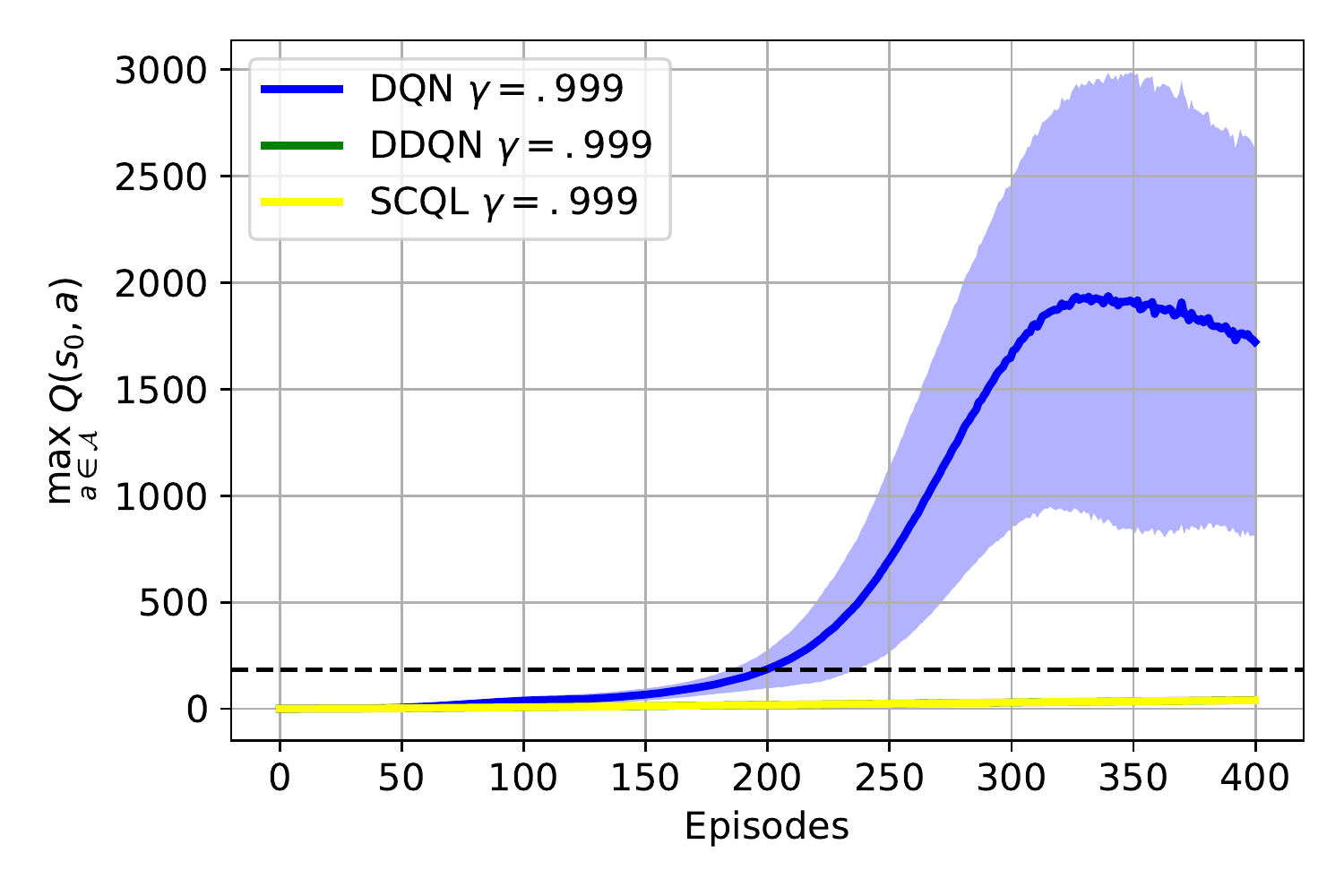}%
  \includegraphics[width=4.8cm,height=\textheight,keepaspectratio]{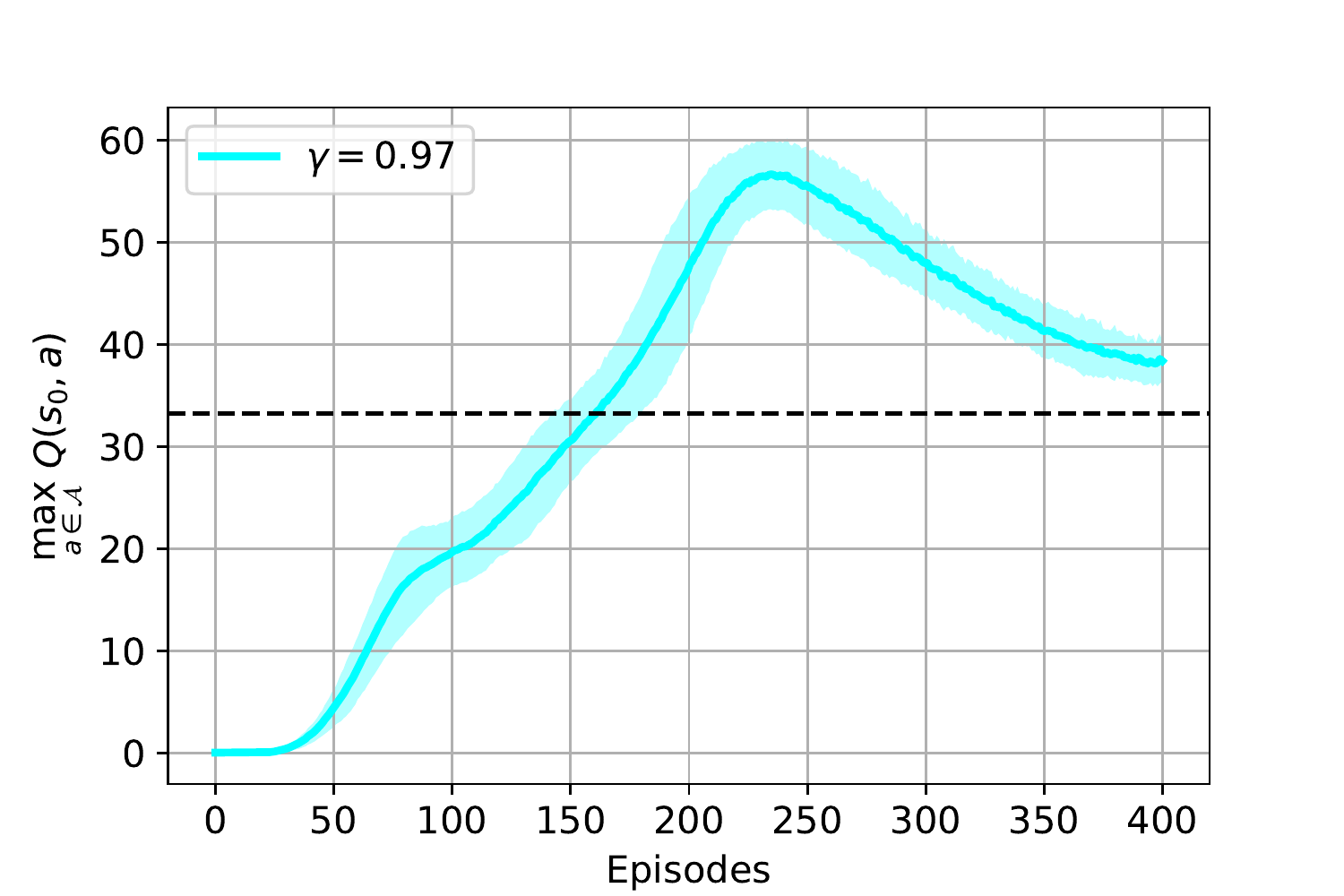}
  
  \includegraphics[width=4.5cm,height=\textheight,keepaspectratio]{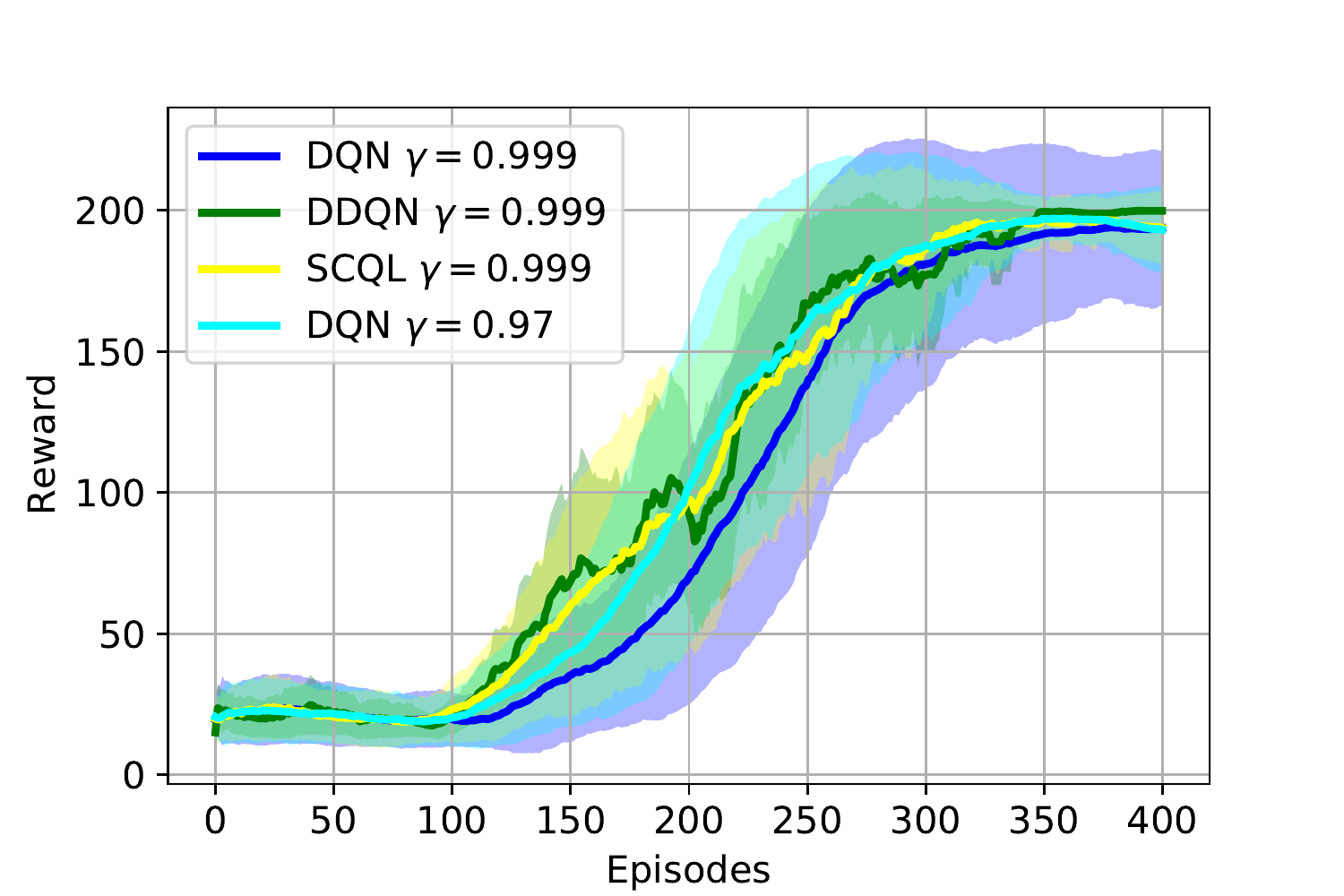}%
\caption{In clockwise order the results showing that: i) DQN suffers from severe overestimation when compared to DDQN and SCQL; ii) training a DQN agent with $\gamma=0.97$ mitigates this overestimation bias significantly; iii) a DQN agent trained with $\gamma=0.97$ performs just as well as all other algorithms trained with $\gamma=0.999$.}
\label{fig:cartpole_results}
\end{figure}

As there are multiple sources of stochasticity in neural networks, it is well known that Deep Q-learning suffers from overestimation even in fully deterministic environments \cite{van2016deep}. We investigate whether this could be prevented by lowering the discount factor. We find that on the \texttt{Cartpole-v0} environment for $\gamma=0.999$, the algorithm performs well but estimations of the starting state (given $\gamma=0.999$ and a maximal cumulative reward of $200$ the optimal Q value is $\sum_{k}^{200} 1 \times \gamma^{k} \approx 181.35$) are highly varied and overly optimistic (see blue line of first plot of Fig.\ref{fig:cartpole_results}). In the same plot we can also observe that Deep Double Q-Learning and Deep Self-Correcting Q-Learning suffer from the same underestimation bias that we already observed in the Gridworld environment (yellow and green lines). However, simply setting $\gamma=0.97$ improves the estimation significantly (second plot of Fig.\ref{fig:cartpole_results}) \footnote{Note that with $\gamma=0.97$ the optimal Q value becomes $\sum_{k}^{200} 1 \times \gamma^{k} \approx 34$.} without being detrimental to the overall performance of the algorithm (see third plot of Fig.\ref{fig:cartpole_results}) which performs on par with a DQN, DDQN and SCQL agent trained with $\gamma=0.999$.  

\begin{table}[ht]
\tiny
\centering
\caption{Our results assessing the degree of overestimation of the different algorithms for all tested environments. We report $\max_{a\in \mathcal{A}}Q(s_0,a)-Q^*(s_0,a)$ where values $<0$ mean  that the algorithms suffer from underestimation bias whereas values $>0$ mean overestimation.}
\input{Tables/absolute_overestimation}
\label{relative-overestimation}

\label{tab:results}
\end{table}

\section{Summary}

We have studied the role of the learning rate $\alpha$, the discount factor $\gamma$ and the reward signal $r$ under the lens of the overestimation bias that characterizes the popular Q-Learning algorithm. Our results suggest that all three factors of influence play a significant role in Q-Learning's estimations. As a result, we have provided initial intuition on how to set such parameters in order to prevent overestimation from happening in the tabular setting as well as in the function approximator one. We believe that stochasticity is among the core problems which causes Q-Learning's $\max$ operator to overestimate, however, if one can work around the impact of stochasticity, as we tried with our methods, one can reduce overestimation without changing the maximization operator. By considering $\alpha, \gamma$ and $\hat{r}$ one can prevent overestimation in an easy, computationally not-intensive way as, differently from methods such as Double Q-Learning and Self-Correcting Q-Learning, there is no need to keep track of a second state-action table. Opportunities for future research are testing the methods on more environments, investigating the effects of an annealing exponential moving average value, and further investigating the applicability of our methods in the Deep Reinforcement Learning regime. Our goal there is to study the effects of $\alpha, \gamma$ and $\hat{r}$ with respect to the \textit{Deadly Triad} of Deep Reinforcement Learning \cite{van2018deepdeadly} and potentially construct a version of the DQN algorithm that does not require the use of a target network.

\bibliography{main}
\bibliographystyle{icml2021}



\onecolumn
\section{Appendix}
\label{sec:appendix}
For the function approximator case we replicate the network architecture and hyperparameters from \citet{chen2021investigation}. Agents are trained for 400 episodes. The exploration strategy is epsilon greedy with an epsilon value that decreases linearly from 1.0 to 0.0. The network is an MLP with a first hidden layer of 48 units and second hidden layer of 96 units. The batch size is 512, and the target network is updated every 64 steps. Adam is used as an optimizer with standard values of $\beta_1 = 0.9$ and $\beta_2 = 0.999$. The learning rate is 0.0001.

All code to replicate the results is available at: \url{https://github.com/overestimationbias/Factors-of-Influence-of-the-Overestimation-Bias}
\end{document}

%% file: Tables/absolute_overestimation.tex
\newcommand*{\MinNumber}{-100.0}%
\newcommand*{\MidNumber}{0.0}%
\newcommand*{\MaxNumber}{100.0}%

\newcommand{\ApplyGradient}[1]{%
  \IfDecimal{#1}{
    \ifdim #1 pt > \MidNumber pt%
    \pgfmathsetmacro{\PercentColor}{max(min(100.0*(#1-\MidNumber)/(\MaxNumber-\MidNumber),100.0),0.0)}%
    \edef\x{\noexpand\cellcolor{\colorpos!\PercentColor!white}}\x\textcolor{black}{#1}%
    \else%
    \pgfmathsetmacro{\PercentColor}{max(min(100.0*(\MidNumber-#1)/(\MidNumber-\MinNumber),100.0),0.0)}%
    \edef\x{\noexpand\cellcolor{\colorneg!\PercentColor!white}}\x\textcolor{black}{#1}%
    \fi%
    }{#1}
}

\newcolumntype{R}{>{\collectcell\ApplyGradient}{c}<{\endcollectcell}}
\begin{tabular}{|l|l|l|l|l|l|l|} 
\cline{2-7}
\multicolumn{1}{l|}{} & QL      & DQL     & SCQL   & $\alpha$ & $\gamma$ & $\hat{r}$  \\ 
\hline
Gridworld             & +14.43  & -7.22  & -6.36  & +0.095   & +1.25    & -1.42      \\ 
\hline
Blackjack             & +0.008  & -0.080 & -0.001 & +0.026   & -        & +0.031     \\ 
\hline
Cartpole              & +1459.3 & -131.9 & -132.3 & -        & +3.47    & -          \\
\hline
\end{tabular}